\documentclass[10pt,twocolumn]{article}

\usepackage{iccv}
\usepackage{times}
\usepackage{epsfig}
\usepackage{graphicx}
\usepackage{amsmath}
\usepackage{amssymb}
\usepackage{multirow}
\usepackage{balance}
\usepackage{subfigure}

\usepackage[pagebackref=true,breaklinks=true,letterpaper=true,colorlinks,bookmarks=false]{hyperref}

\iccvfinalcopy 

\def\iccvPaperID{21} 

\ificcvfinal\fi

\begin{document}

\title{Social and Scene-Aware Trajectory Prediction in Crowded Spaces}

\author{
Matteo Lisotto~~~~~~~~Pasquale Coscia~~~~~~~~Lamberto Ballan\\
Department of Mathematics ``Tullio Levi-Civita'', University of Padova, Italy\\
{\tt\small \{pasquale.coscia,lamberto.ballan\}@unipd.it}
}

\maketitle

\begin{abstract}
Mimicking human ability to forecast future positions or interpret complex interactions in urban scenarios, such as streets, shopping malls or squares, is essential to develop socially compliant robots or self-driving cars. Autonomous systems may gain advantage on anticipating human motion to avoid collisions or to naturally behave alongside people. To foresee plausible trajectories, we construct an LSTM (long short-term memory)-based model considering three fundamental factors: people interactions, past observations in terms of previously crossed areas and semantics of surrounding space. Our model encompasses several pooling mechanisms to join the above elements defining multiple tensors, namely social, navigation and semantic tensors. The network is tested in unstructured environments where complex paths emerge according to both internal (intentions) and external (other people, not accessible areas) motivations. As demonstrated, modeling paths unaware of social interactions or context information, is insufficient to correctly predict future positions. Experimental results corroborate the effectiveness of the proposed framework in comparison to LSTM-based models for human path prediction.
\end{abstract}

\section{Introduction}
\label{sec:intro}

Human trajectory forecasting is a relevant topic in computer vision due to numerous applications which could benefit from it. Socially-aware robots need to anticipate human motion in order to optimize their paths and to better comply with human motion. Delivery robots could reduce energy consumption to get to their destinations avoiding people and obstacles as well. Moreover, anomalous behaviors could be detected using fixed cameras in urban open spaces (e.g., parks, streets, shopping malls, etc.) but also in crowded areas (e.g., airports, railway stations). Despite meaningful results attained using recurrent neural networks for time-series prediction, many problems still remain.
In this context, data-driven approaches are usually unaware of surrounding elements which represent one of the main reasons of direction changes in a urban scenario.

\begin{figure}[!t]
\center
\includegraphics[width=1\columnwidth]{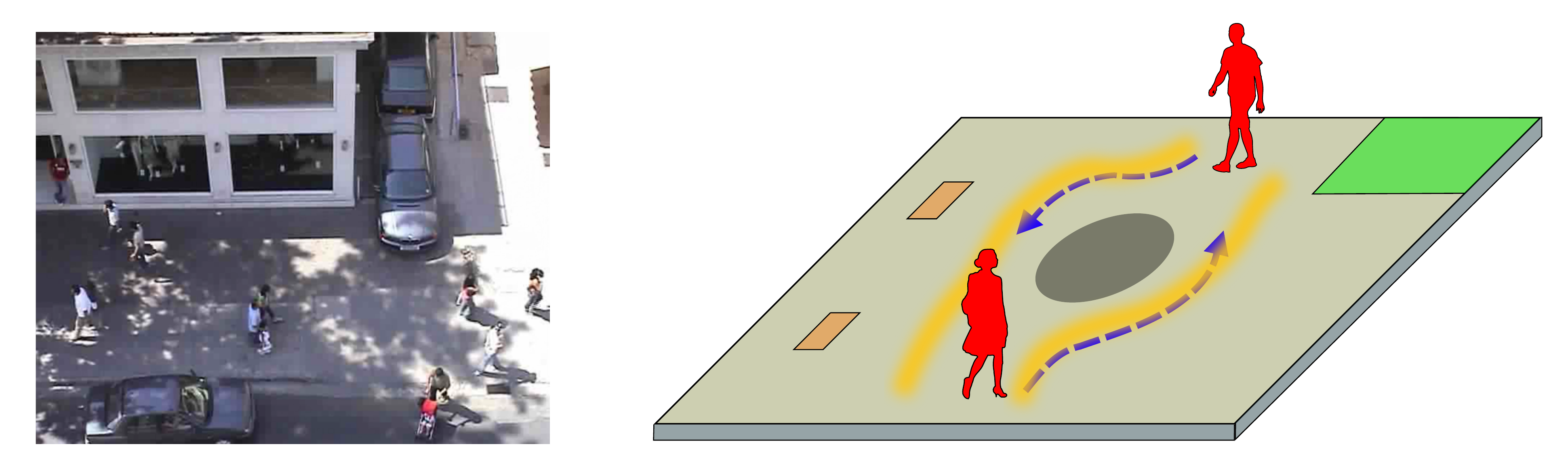}
\caption{Our goal is to predict future positions of pedestrians in an urban scenario. Since human motion is guided by intentions, experience and the surrounding environment, such elements are encapsulated in our framework along with learned social rules to forecast a social and semantic compliant motion in the crowd.}
\label{fig:1}
\end{figure}

When approaching their destination, people tend to conform to observed patterns coming from experience and visual stimuli to avoid threats or select the shortest route. Moreover, when walking in public spaces, they typically take into account which kind of \textit{objects} encounter in their neighborhood. Several factors may also lead to velocity direction changes in many situations. For example, when approaching roundabouts (see Fig.~\ref{fig:1}), people adjust their path to avoid collisions. In some cases, they use different paces according to weather conditions or crowded areas.
In this context, LSTM networks have been extensively used over the last years due to their ability to learn, remember and forget dependencies through gates~\cite{journals/corr/KarpathyJL15,FERNANDO2018466}.  Such characteristics have made them one of the most suitable solution for solving sequence to sequence problems.
To address the limitations of previous works which mainly focuses on modelling \emph{human-human} interactions~\cite{Alahi_2016_CVPR,Hasan2018MXLSTMMT,Gupta2018SocialGS}, we propose a comprehensive framework for predicting future positions which are also locally-aware of surrounding space combining social and semantic elements. Scene context information is crucial to improve prediction of future positions adding physical constraints and providing more realistic paths, as demonstrated by early works focused on exploiting \emph{human-space} interactions~\cite{kitani2012activity,kooij2014context,Ballan_2016_ECCV}.

In this paper, we propose a data-driven approach allowing an LSTM-based architecture to extract social conventions from observed trajectories and augment such data with semantic information of the neighborhood. More specifically, our work is built upon the {\textit Social}-LSTM model proposed by Alahi \etal~\cite{Alahi_2016_CVPR}, in which the network is only aware of nearby pedestrians, by embedding new factors encoding also \emph{human-space} interactions in order to attain more accurate predictions. More specifically, we encompass prior motion about the scene as a navigation map which embodies most frequently crossed areas and scene context using semantic segmentation to restrain motion to more plausible paths.

The remainder of the paper is organized as follows. Section~\ref{sec:relwork} reviews main work related to human path prediction. Section~\ref{sec:semantic-lstm} describes the proposed model. Section~\ref{sec:experiments} provides our findings while conclusions and suggestions for future work are summarized in Section~\ref{sec:conclusion}.

\section{Related Work}
\label{sec:relwork}

We briefly review main work on human path prediction considering two main kinds of interactions, namely \textit{human-human} and \textit{human-human-space}. The former only models interactions among pedestrians; the latter, takes also into account interactions with surrounding elements, i.e. fixed obstacles, which kind of area is crossed (\eg, sidewalk, road) and nearby space.

\paragraph{Human-human interactions.}
Helbing and Moln\'ar~\cite{PhysRevE.51.4282} introduced the Social Force Model to describe social interactions among people in crowded scenarios using hand-crafted functions to form coupled Langevin equations. More recent works based on LSTM networks mainly rely on the model proposed in~\cite{Alahi_2016_CVPR} where a ``social'' pooling mechanism allows pedestrians to share their hidden representations. The key idea is to merge hidden states of nearby pedestrians to make each trajectory aware of its neighbourhood. Nevertheless, pedestrians are unaware of nearby elements, such as benches or trees, which could be primary reasons for direction changing when they do not interact with each others. \cite{10.1007/978-3-030-11015-4_18} detects groups of people moving coherently in a given direction which are excluded from the pooling mechanism. \cite{Gupta2018SocialGS} uses a Generative Adversarial Network (GAN) to discriminate between multiple plausible paths due to the inherently multi-modal nature of trajectory forecasting task. The pooling mechanism relies on relative positions between two pedestrians. The model captures different styles of navigation but does not make any differences between structured and unstructured environments. \cite{Vemula2018SocialAM} handles prediction using a spatio-temporal graph which models both position evolution and interaction between pedestrians. \cite{Hasan2018MXLSTMMT} embodies \emph{vislet} information within the social-pooling mechanism also relying on mutual faces orientation to augment space perception.

\paragraph{Human-human-space interactions.}
Sadeghian \etal \cite{Sadeghian2018SoPhieAA} adopt a similar approach to ours, by taking into account both past crossed areas and semantic context to predict social and context-aware positions using a GAN.
\cite{Bartoli2018ContextAwareTP} introduces attractions towards static objects, such as artworks, which deflect straight paths in several scenarios (\eg{, museums, galleries}) but the approach is limited to a reduced number of static objects. \cite{Ballan_2016_ECCV} proposes a Bayesian framework based on previously observed motions to infer unobserved paths and for transferring learned motions to unobserved scenes. Similarly, in \cite{COSCIA201881} circular distributions model dynamics and semantics for long-term trajectory predictions. \cite{Sadeghian_2018_ECCV} uses past observations along with bird's eye view images based on a two-levels attention mechanism. The work mainly focuses on scene cues partially addressing agents' interactions.

Some relevant approaches do not fall into the above two categories. For example,  \cite{Shen2018TransferablePM} focuses on transfer learning for pedestrian motion at intersections using Inverse Reinforcement Learning (IRL) where paths are inferred exploiting goal locations. \cite{10.1007/978-3-030-11015-4_13} attains best performance on the challenging Stanford Drone Dataset (SDD)~\cite{sadeghiankosaraju2018trajnet} using a recurrent-encoder and a dense layer. \cite{Dark_Matter} predicts future positions in order to satisfy specific needs and to reach latent sources.

\begin{figure*}
\center
\includegraphics[width=0.7\linewidth]{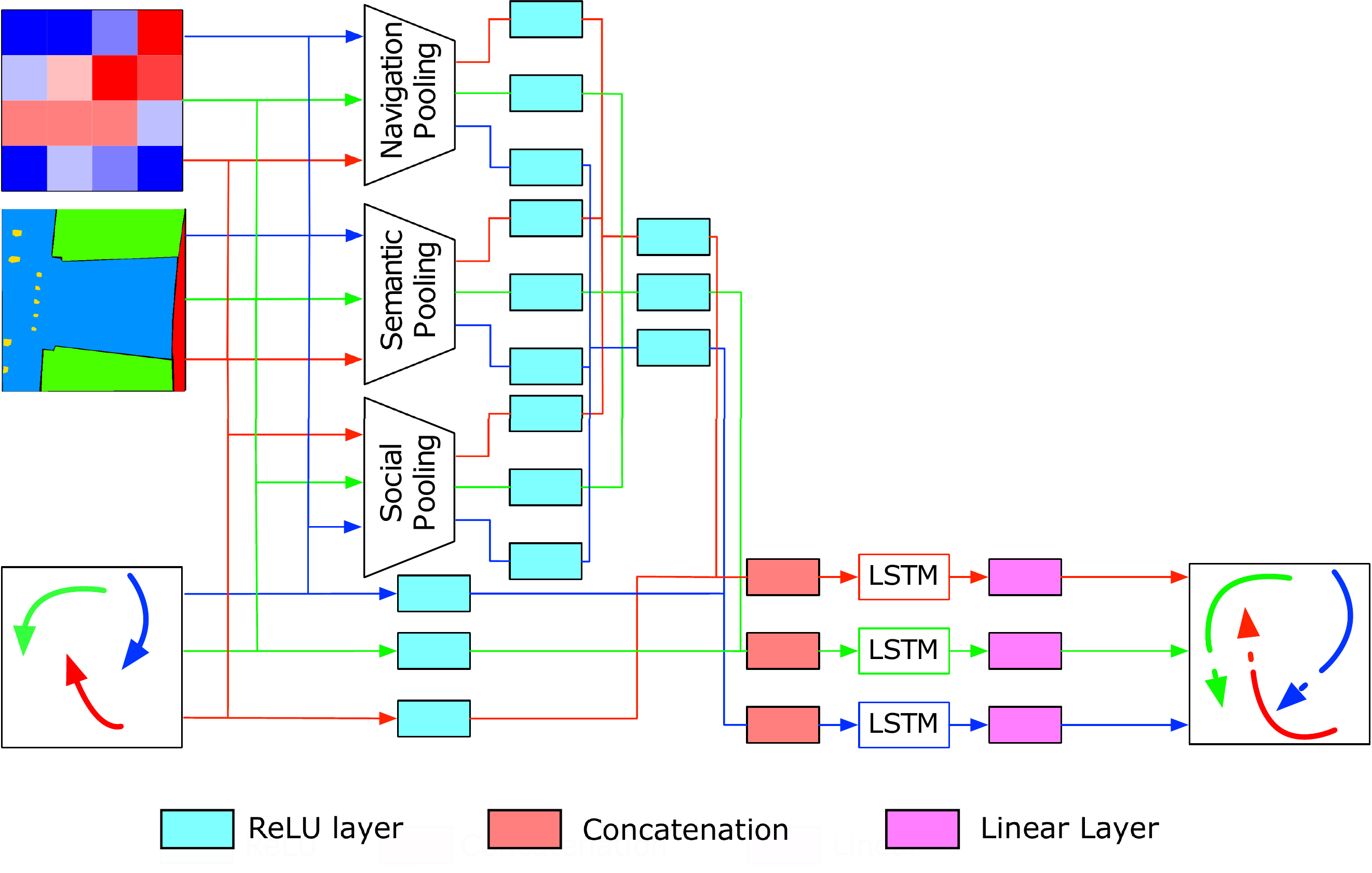}
\caption{Overview of the proposed model. Trajectories, navigation map and semantic image are fed to the LSTM network and combined using three pooling mechanisms, namely social, navigation and semantic pooling. Future positions are obtained using linear layers to extract key parameters of a Gaussian distribution.}
\label{fig:model}
\end{figure*}

\section{Our model}
\label{sec:semantic-lstm}

Pedestrian dynamics in urban scenarios are highly influenced by static and dynamic factors which guide people towards their destinations. For example, grass is typically less likely to be crossed than sidewalks or streets for pedestrians. Benches are turned around by people walking with accelerated paces. Moreover, only doors are used to enter buildings. Hence, to forecast realistic paths, it is important to allow human dynamics to be influenced by surrounding space, not only in terms of other people in their neighborhood, but also considering semantics of crossed areas as well as past observations which can represent our experience. To this aim, we extend the Social-LSTM model proposed in~\cite{Alahi_2016_CVPR}, as schematized in Fig.~\ref{fig:model}. More specifically, our framework models each pedestrian as an LSTM network interacting with the surrounding space using three pooling mechanisms, namely \textit {Social, Navigation} and \textit{Semantic} pooling.
\textit{Social} pooling mechanism takes into account the neighborhood in terms of other people, merging their hidden states. \textit{Navigation} pooling mechanism exploits past observations to discriminate between equally likely predicted positions using previous information about the scene. Finally, \textit{Semantic} pooling uses semantic scene segmentation to recognize not crossable areas. 
\vspace{1em}

Given the $i^{th}$ pedestrian, his/her complete trajectory is represented by the 2-D sequence $\mathcal{T}^i = \{(x_1^i, y_1^i), (x_2^i, y_2^i), ..., (x_{T_{obs}}^i, y_{T_{obs}}^i), (x_{T_{obs}+1}^i, y_{T_{obs}+1}^i), ...,\\ (x_{T_{pred}}^i, y_{T_{pred}}^i)\}$ where $T_{obs}$ and $T_{pred}$ represent the last observation and prediction timestamps, respectively. Each trajectory is then associated to an LSTM network which is described by the following equations:

\begin{equation*}
{\bf f}^i_t = \sigma(W_f {\bf x}^i_t + U_f {\bf h}^i_{t-1} + b_f)
\end{equation*}
\begin{equation*}
{\bf i}^i_t = \sigma(W_i {\bf x}^i_t + U_i {\bf h}^i_{t-1} + b_i)
\end{equation*}\vspace{-1.2em}
\begin{equation*}
{\bf o}^i_t = \sigma(W_o {\bf x}^i_t + U_o {\bf h}^i_{t-1} + b_o)
\end{equation*}
\begin{equation*}
{\bf c}^i_t = {\bf f}^i_t \odot {\bf c}^i_{t-1} + {\bf i}^i_t \odot tanh(W_c {\bf x}^i_t + U_c {\bf h}^i_{t-1} + b_c)
\end{equation*}\vspace{-1.25em}
\begin{equation}
{\bf h}^i_t = {\bf o}^i_t \odot tanh({\bf c}^i_t)
\end{equation}
where ${\bf f}^i_t$, ${\bf i}^i_t$, ${\bf o}^i_t$ are the forget, input and output gates, respectively; ${\bf c}^i_t$ is the cell state and ${\bf h}^i_t$ is the hidden state. $\odot$ indicates the element-wise product.

The above model, named \textit{Vanilla} LSTM, is unaware of what happens nearby the monitored agent, such as the presence of other people, encountered obstacles and most frequently crossed areas. In fact, such a network could only learn motion patterns and dependencies potentially present in the training set's trajectories. To consider a more rich input representation, \textit{Vanilla} LSTM is enhanced concatenating the following tensors:

\paragraph{Social Tensor.}
As proposed in~\cite{Alahi_2016_CVPR}, we firstly exploit a social pooling mechanism in order to make people aware of their neighbors. More specifically, hidden states of people in the neighborhood are taken into account using an $N_o \times N_o \times D$ social tensor:
\begin{equation}
H^i_t(m, n, :) = \sum_{j \in \mathcal{N}_i} {\bf 1}_{mn}[x^j_t-x^i_t, y^j_t-y^i_t] {h^j}_{t-1}
\end{equation}
where $N_o$ is neighborhood size and $D$ is the dimension of the hidden state. The indicator function ${\bf 1}_{mn}(x,y)$ checks if (x,y) is inside the $(m,n)$ cell.  

\begin{figure}
\centering
\includegraphics[width=0.95\linewidth]{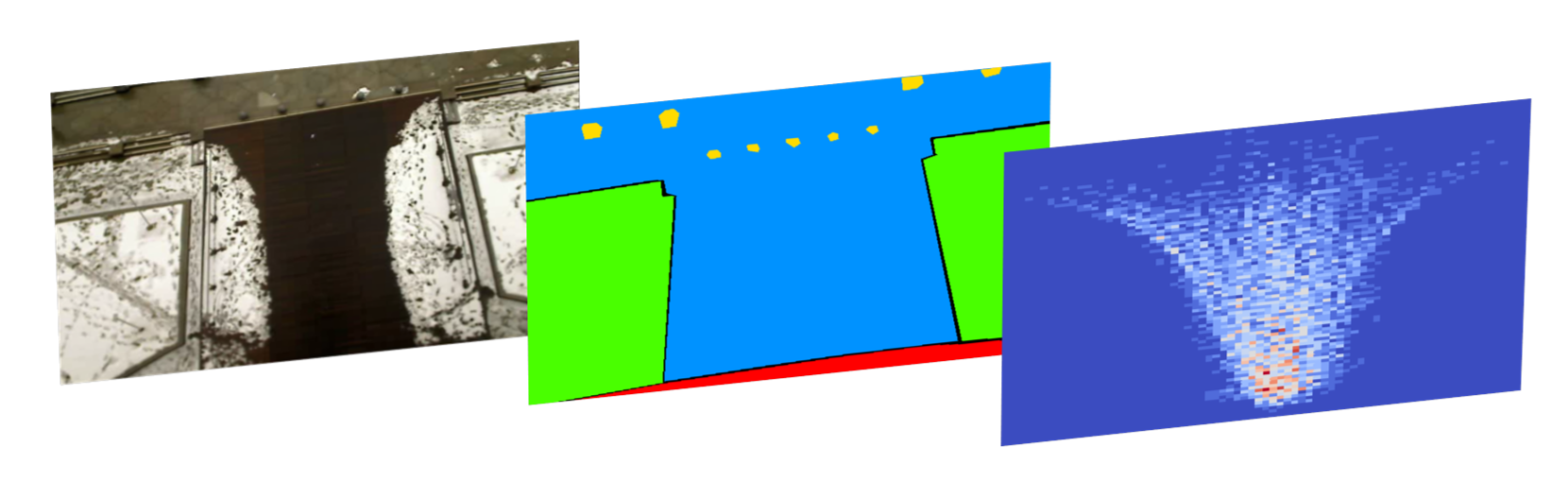}
\caption{Semantic map is generated from the reference image while the Navigation map is obtained from observed data. The image shows an example of such maps for ETH dataset.}
\label{fig:eth}
\end{figure}

\paragraph{Navigation Tensor.}
People tend to reach building entrances or opposite sidewalks using a limited number of paths. Some areas would, indeed, be more likely to be crossed than others. On the contrary, areas corresponding to obstacles or buildings would not (or less likely to) be crossed making them not \textit{eligible} for generating new position candidates. To measure such crossing probability, we define the \textit{Navigation Map} $\mathcal{N}$ which counts the crossing frequency of squared patches. A smoothing linear filter (i.e., average pooling) is then used to reduce ``sharp'' frequency transitions. An example of such map in shown in Fig.~\ref{fig:eth}.
Given the \textit{Navigation Map} $\mathcal{N}$, we define the rank-2 \textit{Navigation} tensor $N^i_t \in N_n \times N_n$ as follows:
\begin{equation}
    N^i_t(m,n) = \mathcal{N}_{mn},
\end{equation}
which extracts the neighborhood's frequency of the $i^{th}$ pedestrian for the $(m,n)$ cell considering all the past observations for such cell.

\paragraph{Semantic Tensor.}
People may also manifest direction changes due to a number of reasons; for example, they could be approaching a fixed obstacle which must be circled or could avoid streets preferring sidewalks. The aim of the \textit{Semantic} tensor is to capture \textit{why} specific dynamics emerge related to the semantics of surrounding space. Since our datasets do not provide any semantic annotations to model human-space interactions, we define the following semantic classes $\mathcal{C}$ = \{grass, building, obstacle, bench, car, road, sidewalk\}. A one-hot encoding is used to represent semantic of pixels image. For example, assuming that a pixel represents grass, a location $j$ is represented by a vector ${\bf v}_j \in \mathbb{R}^7 = [1~0~...~0]$ according to $\mathcal{C}$.\\Given a neighborhood size of $N_s$, we define a $N_s \times N_s \times L$ tensor $S^i_t$ for the $i^{th}$ pedestrian as follows:
\begin{equation}
S^i_t(m, n, :) = \frac{1}{|\mathcal{S}_{mn}|} \sum_{j \in \mathcal{S}_{mn}} {\bf{v}}_j
\end{equation}
where ${\bf{v}}_j$ represents the semantic vector of location $j$ and $\mathcal{S}_{mn}$ represents the locations within the (m,n) cell of $i^{th}$ pedestrian. $|\mathcal{S}_{mn}|$ is the number of locations within the $(m,n)$ cell. In other words, for each cell, we extract the occurring frequency of each semantic class.
\vspace{1em}

\begin{figure}
\center
\includegraphics[width=0.95\linewidth]{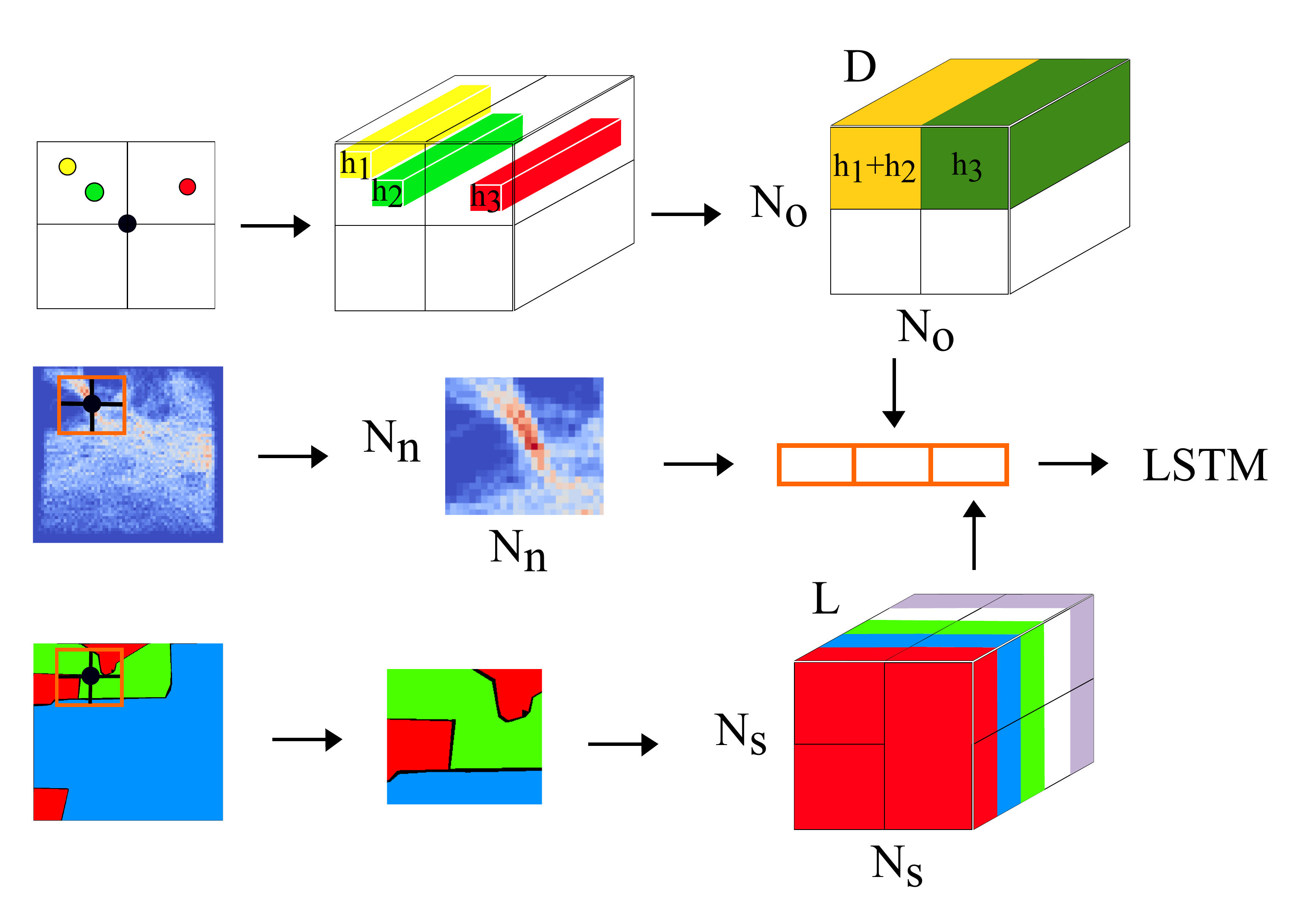}
\caption{Overview of the pooling mechanisms. Three tensors take into account social neighborhood, past observations and semantics of surrounding space, respectively. Tensors are finally concatenated, processed by ReLU layers and fed to LSTM networks along with embedded positions. Figure also highlights dimensions of each introduced tensor.}
\label{fig:pooling}
\end{figure}

The above tensors are embedded into three vectors, namely $a^i_t, n^i_t, s^i_t$ while the spatial coordinates into $e^i_t$. The embedded vectors are concatenated and used as input to the LSTM cell as follows:

\begin{equation*}
    e^i_t = \Phi(x^i_t, y^i_t; W_e)
\end{equation*}
\begin{equation*}
    a^i_t = \Phi(H^i_t; W_a)
\end{equation*}
\begin{equation*}
    n^i_t = \Phi(N^i_t; W_n)
\end{equation*}
\begin{equation*}
    s^i_t = \Phi(S^i_t; W_s)
\end{equation*}
\begin{equation*}
    g^i_t = \Phi(concat(a^i_t, n^i_t, s^i_t); W_g)
\end{equation*} \vspace{-1.25em}
\begin{equation}
    h^i_t = LSTM(h^i_{t-1}, concat(e^i_t, g^i_t); W_h)
\end{equation}
where $\Phi$ represents the ReLU activation function and $W_h$ are the LSTM weights. Fig.~\ref{fig:pooling} depicts our pooling mechanisms.

\paragraph{Loss Function.}
Positions are predicted using a bi-variate Gaussian distribution whose parameters are obtained using a $D \times 5$ linear layer as follows:
\begin{equation}
    [\mu^i_t, \sigma^i_t, \rho^i_t] = W_lh^i_{t-1},
\end{equation}\vspace{-1.25em}
\begin{equation}
    (\hat{x}^i_t, \hat{y}^i_t) \sim \mathcal{N}((x,y); \mu^i_t, \sigma^i_t, \rho^i_t).  
\end{equation}
Finally, the parameters of the network are obtained minimizing the negative log-Likelihood loss $L^i$ for the $i^{th}$ pedestrian as follows:
\begin{multline}
    L^i(W_e, W_a, W_n, W_s, W_g, W_h, W_l) =\\ -\sum^{T_{pred}}_{t = T_{obs}+1} log(\mathbb{P}(x^i_t, y^i_t|\mu^i_t, \sigma^i_t, \rho^i_t)).    
\end{multline}
The above loss is minimized for all the trajectories in our training sets.

\section{Experiments}
\label{sec:experiments}
In this section, we describe the used datasets along with the evaluation protocol. Next, we present a quantitative analysis to show the effectiveness of our model. Finally, we show some qualitative results of predicted trajectories for challenging situations.

\begin{table*}[!t]
\centering
\resizebox{0.85\linewidth}{!}{\begin{tabular}{|c|c|c|c|c|c|c|}
\hline
Metric & Scene & Vanilla LSTM & Social-LSTM~\cite{Alahi_2016_CVPR} & SN-LSTM & SS-LSTM & SNS-LSTM\\
\hline\hline
{\multirow{5}{*}{ADE}} & ETH~\cite{Pellegrini2009YoullNW} & $0.52$ & $0.51$ & {\bf 0.47} & $0.48$ & $0.58$\\
{} & HOTEL~\cite{Pellegrini2009YoullNW} & $0.33$ & $0.31$ & $0.44$ & {\bf 0.24} & $0.30$\\
{} & UNIV~\cite{Lerner2007CrowdsBE} & $0.52$ & $0.55$ & $0.39$ & $0.43$ & {\bf 0.37}\\
{} & ZARA-01~\cite{Lerner2007CrowdsBE} & $0.41$ & $0.36$ & $0.29$ & $0.33$ & {\bf 0.28}\\
{} & ZARA-02~\cite{Lerner2007CrowdsBE} & $0.27$ & {\bf 0.25} & $0.28$ & $0.31$ & $0.26$\\
\hline
{} & {\bf Average} & $0.41 \pm 0.11$ & $0.40 \pm 0.13$ & $0.37 \pm 0.09$ & ${\bf 0.36 \pm 0.10}$ & $0.36 \pm 0.13$\\
\hline
{\multirow{5}{*}{FDE}} & ETH~\cite{Pellegrini2009YoullNW} & $2.84$ & $2.82$ & $2.55$ & $2.57$ & {\bf2.43}\\
{} & HOTEL~\cite{Pellegrini2009YoullNW} & $1.90$ & $1.67$ & $2.25$ & {\bf1.38} & $1.58$\\
{} & UNIV~\cite{Lerner2007CrowdsBE} & $2.92$ & $3.04$ & $2.10$ & $2.54$ & {\bf2.08}\\
{} & ZARA-01~\cite{Lerner2007CrowdsBE} & $2.35$ & $2.05$  & $1.56$ & $1.81$ & {\bf1.53}\\
{} & ZARA-02~\cite{Lerner2007CrowdsBE} & $1.48$ & {\bf1.42} &  $1.59$ & $1.63$ & $1.44$\\
\hline
{} & {\bf Average} & $2.30 \pm 0.61$ & $2.20 \pm 0.71$ &  $2.01 \pm 0.43$ & $1.99 \pm 0.54$ & {\bf1.81 $\pm$ 0.43}\\
\hline
\end{tabular}
}\vspace{5pt}
\caption{Quantitative results of our architecture compared to baseline models for ETH and UCY datasets. The errors are reported in meters. Our models attain on average best performance due to the combination of learned social rules, past observations and acquired information of the surrounding space. For the ADE metric our two models, namely SS-LSTM and SNS-LSTM, show comparable results. By contrast, the best FDE value is attained considering the combination of all factors.}\label{tab:ethucy}
\end{table*}

\paragraph{Datasets.}
For our experiments, we use two datasets: ETH~\cite{Pellegrini2009YoullNW} and UCY~\cite{Lerner2007CrowdsBE}. ETH contains two scenes ({\itshape ETH} and {\itshape HOTEL}) while UCY contains three scenes ({\itshape UNIV/UCY}, {\itshape ZARA-01}, {\itshape ZARA-02}). They are captured from a bird's-eye view and involve numerous challenging situations, such as interacting pedestrians, standing people and highly non-linear trajectories.  We use a leave-one-out-cross-validation approach training the models on N-1 scenes and testing on the remaining one. We average the results over the five datasets.

\paragraph{Evaluation Protocol.}
To perform an ablation study, we separately test the effect of navigation and semantic factors both along with social interactions, defining two different models, namely SN-LSTM and SS-LSTM, respectively. Finally, we combine all the above effects defining the SNS-LSTM model.
As proposed in~\cite{Pellegrini2009YoullNW}, a pedestrian trajectory is observed for $3.2 s$ in order to predict the next $4.8 s$. At frame level, we train the network on $8$ frames and predict the next $12$ frames. As error metrics, we report the \textit{Average Displacement Error} (ADE) and the \textit{ Final Displacement Error} (FDE). ADE represents the average Euclidean distance between the predicted and ground-truth positions, while FDE consists in the average Euclidean distance between the final predicted and ground-truth position. The above metrics are defined as follows:
\begin{equation}
    ADE = \frac{\sum_{i \in \mathcal{P}} \sum^{T_{pred}}_{t = T_{obs}+1}\sqrt{((\hat{x}^i_t, \hat{y}^i_t)-(x_t^i, y_t^i))^2}} {|\mathcal{P}|\cdot T_{pred}},
\end{equation}
\begin{equation}
    FDE = \frac{\sum_{i \in \mathcal{P}} \sqrt{((\hat{x}^i_{T_{pred}}, \hat{y}^i_{{T_{pred}}})-({x_{{T_{pred}}}}^i, {y_{{T_{pred}}}}^i))^2}} {|\mathcal{P}|},
\end{equation}

where $\mathcal{P}$ is the set of pedestrians and $|\mathcal{P}|$ its cardinality, $(\hat{x}^i_t, \hat{y}^i_t)$ are the predicted coordinates at time $t$ and $(x_t^i, y_t^i)$ are the groud-truth coordinates at time $t$. \\We compare our models against two LSTM-based methods, i.e., Vanilla LSTM and Social-LSTM~\cite{Alahi_2016_CVPR}.

\paragraph{Implementation Details.}
The number of hidden units for each LSTM cell and the embedding dimension of spatial coordinates are set to 128 and 64, respectively. The model is trained on a single GPU using TensorFlow library\footnote{The code is released at \url{https://github.com/Oghma/sns-lstm/}.}
. The learning rate is set to 0.003 and we use RMS-prop as optimizer with a decay of 0.95. Models are trained for 50 epochs.  $N_o, N_n,$ and $N_s$ are set to 8, 32 and 20, respectively.

\subsection{Quantitative Results}
Table~\ref{tab:ethucy} shows quantitative results for ETH and UCY datasets. As expected, the worst model is the one which does not take into account any internal/external factor, namely Vanilla-LSTM.  The lack of any kind of interactions does not allow the model to reproduce realistic paths. We also notice that SN-LSTM model improves the performance compared to S-LSTM model due to the introduction of discriminative regions especially when two or more path are plausible. A significant improvement is also obtained when semantics is introduced with SS-LSTM, especially for HOTEL scene. Interestingly, S-LSTM performs better on ZARA-02 scene, where results are slightly better compared to SNS-LSTM. The main reason could be ascribed to a number of non-moving pedestrians of such dataset where navigation and semantic factors may not much influence the prediction. Navigation map allows better predicted trajectories on ETH dataset compared to HOTEL dataset. For the latter, the effect of only semantic pooling mechanism reduces the error metrics of $\sim50\%$. However, the combination of our proposed factor seems to confirm the importance of introducing navigation and semantic factors to achieve more robust predictions.

\subsection{Qualitative Results}

To perform a qualitative study, we firstly show static comparisons between our models and baselines for some trajectories and then show several predicted trajectories over time evaluating predictions at successive timestamps.

\begin{figure}
\center \includegraphics[height=0.85cm, width=7.9cm]{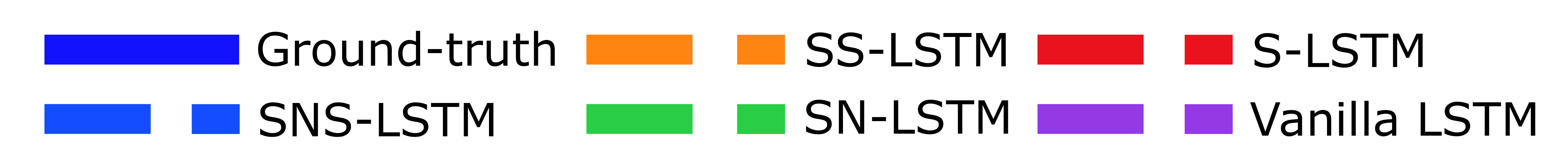}\\
        \includegraphics[height=3cm, width=4cm]{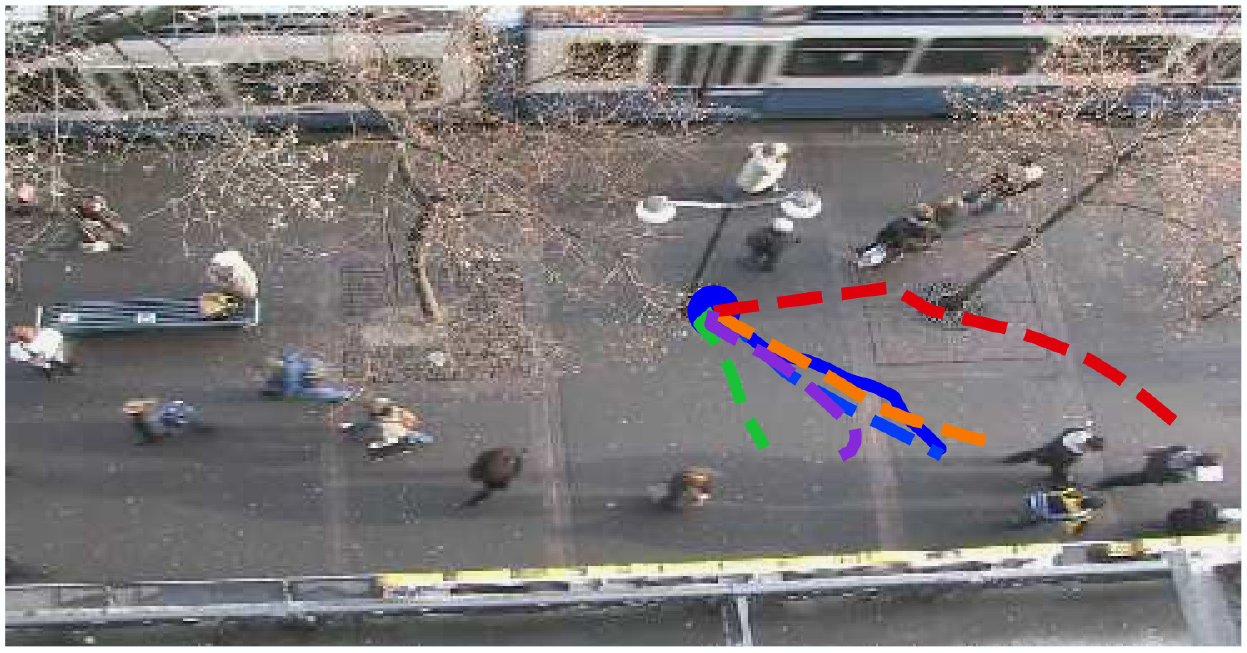}
        \label{fig:first_sub}
        \includegraphics[height=3cm, width=4cm]{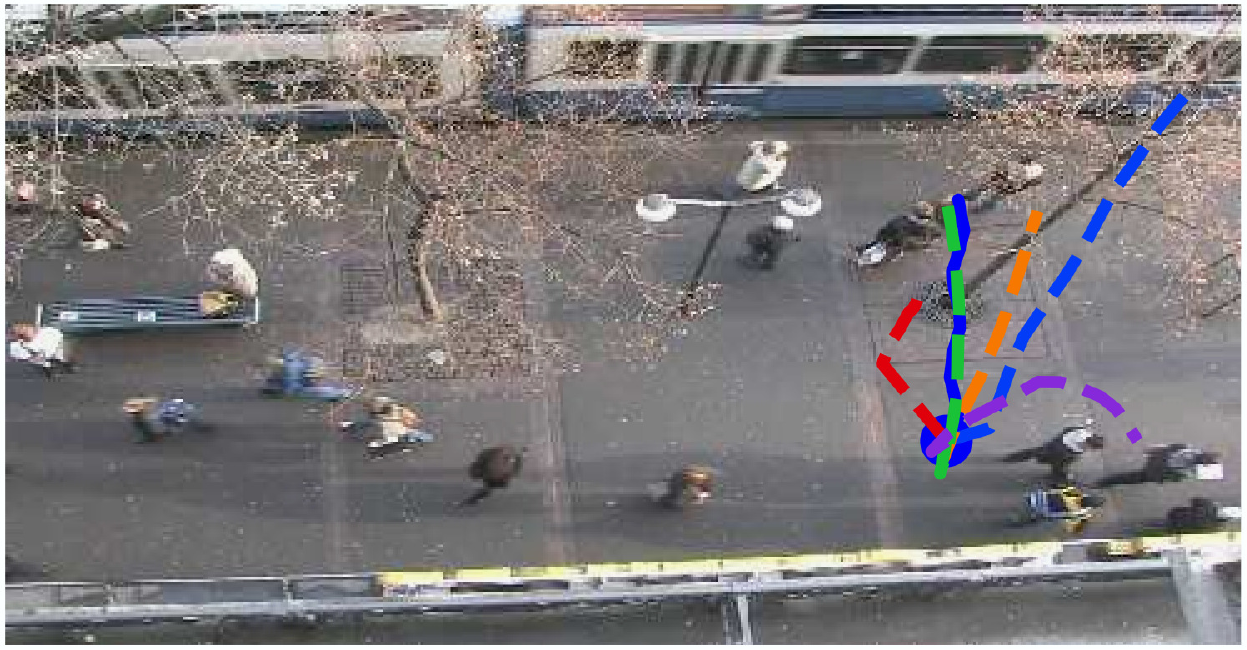}
        \label{fig:fourth_sub}
        
        \includegraphics[height=3cm, width=4cm]{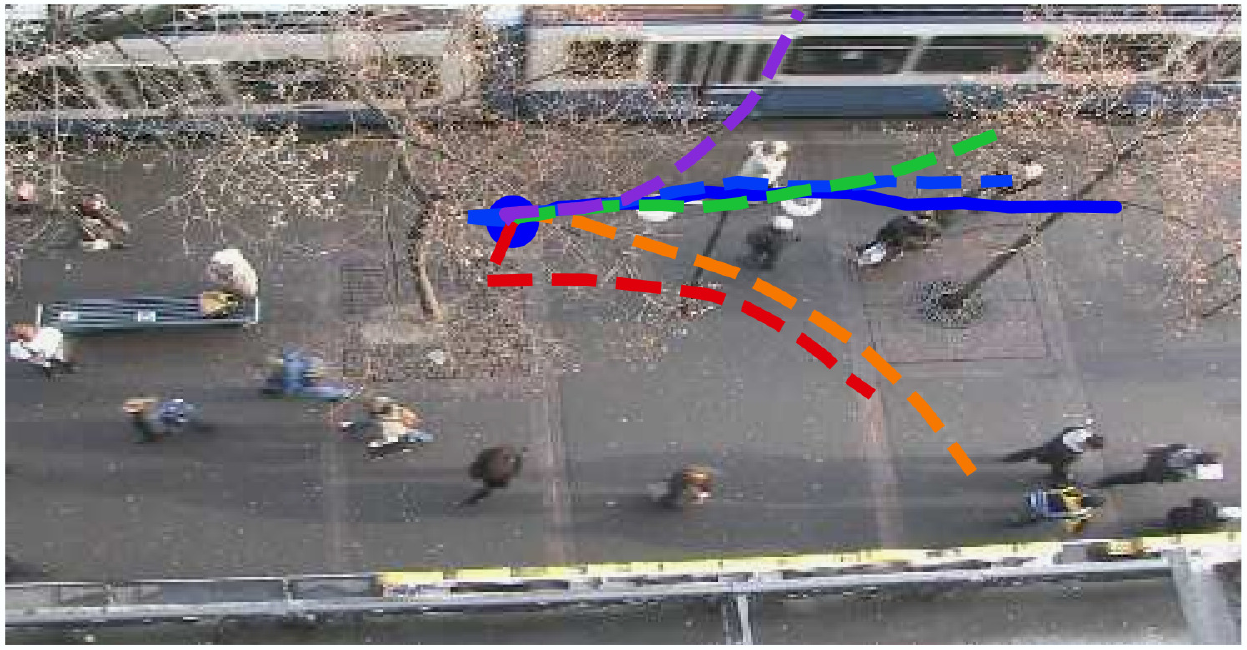}
        \label{fig:second_sub}
        \includegraphics[height=3cm, width=4cm]{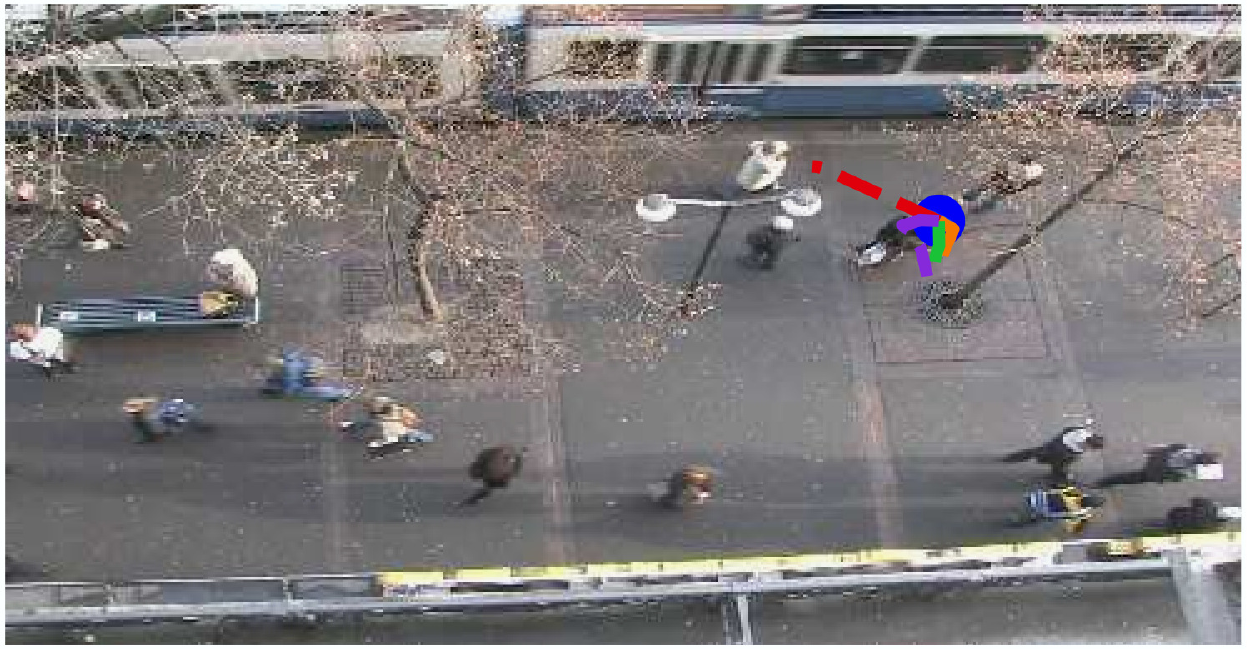}
        \label{fig:fifth_sub}

        \includegraphics[height=3cm, width=4cm]{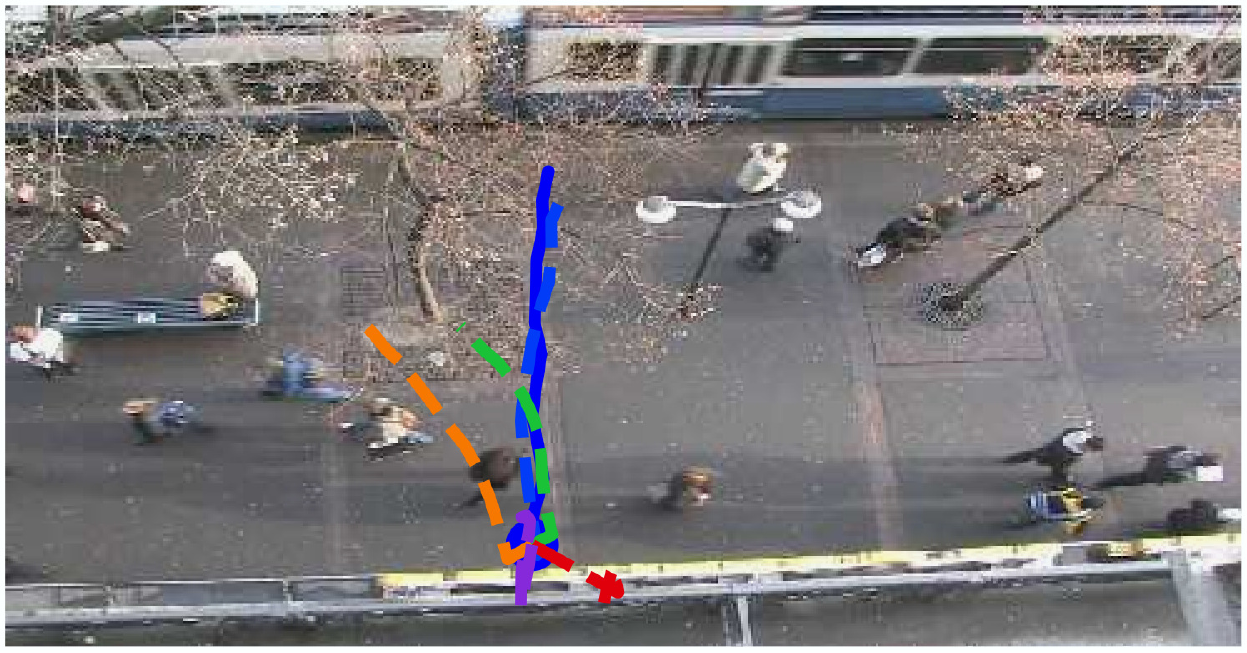}
        \label{fig:third_sub}
        \includegraphics[height=3cm, width=4cm]{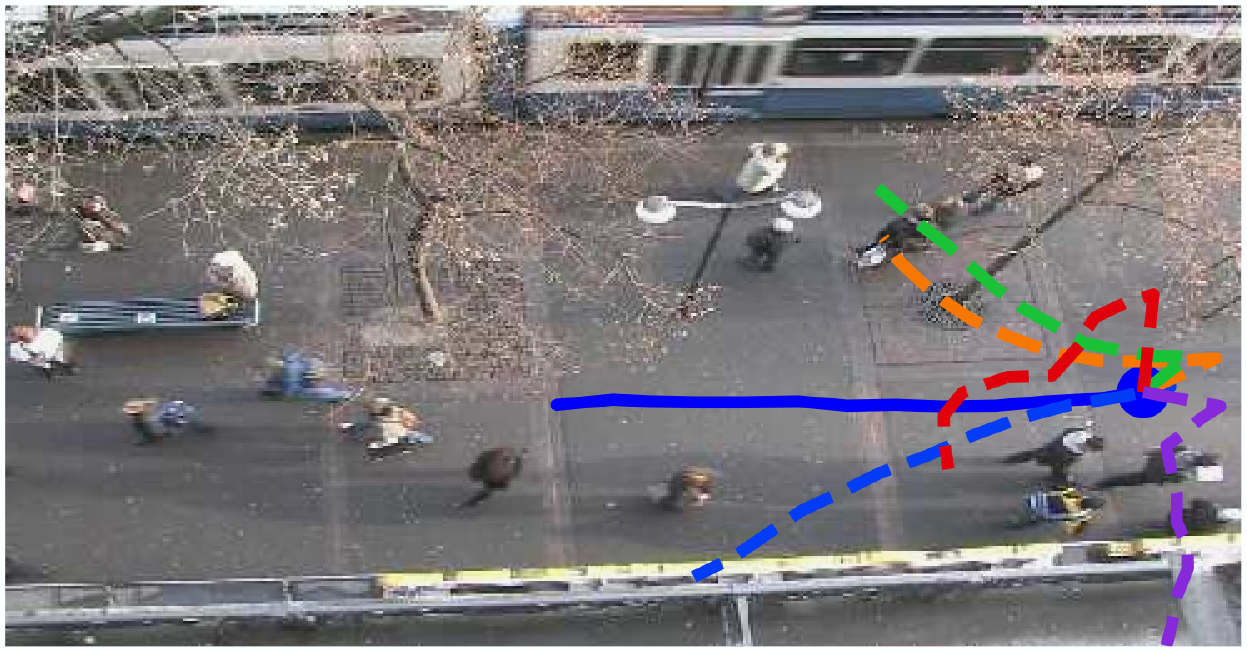}
        \label{fig:sixth_sub}

    \vspace{5pt}
    \caption{Some examples of predicted trajectories for HOTEL dataset. Ground-truths are shown as solid lines, while predicted trajectories as dashed lines. First column shows cases where predicted positions are very close to the real paths. Second column shows cases where the SNS-LSTM appears not able to correctly predict future positions.}
    \label{fig:qual_subfigures_HOTEL}
\end{figure}

\paragraph{Static visualization.} Fig.~\ref{fig:qual_subfigures_HOTEL} shows some examples of predicted trajectories drawn from HOTEL dataset. More specifically, the first column shows cases when our models are able to correctly predict the ground-truth both when people move through the crowd or when they approach the tram. The predicted paths of our models appear able to better capture complex dynamics showing more \textit{natural} motions without continuously adjusting moving directions. We also note that our models are closer to the ground-truth while the baselines end first. Since SN-LSTM and SNS-LSTM are typically similar, our models rely on the navigation map when multiple trajectories can be considered. On the contrary, second column shows cases where SNS-LSTM model appears unable to capture the correct path reaching different destinations. A challenging situation where a standing pedestrian is captured is also shown. In such case, most of the models do not move from the initial position.

\begin{figure*}
\centering     
\subfigure{\label{fig:a11}\includegraphics[width=40mm, height=35mm]{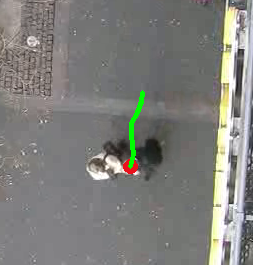}}
\subfigure{\label{fig:a12}\includegraphics[width=40mm, height=35mm]{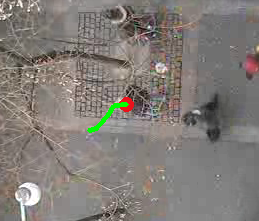}}
\subfigure{\label{fig:a13}\includegraphics[width=40mm, height=35mm]{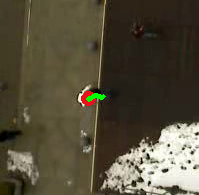}}
\subfigure{\label{fig:a14}\includegraphics[width=40mm, height=35mm]{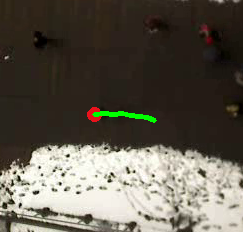}}\\
\vspace{1em}
\subfigure{\label{fig:a21}\includegraphics[width=40mm, height=35mm]{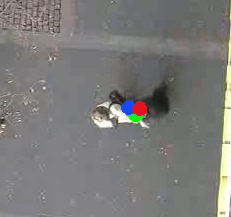}}
\subfigure{\label{fig:a22}\includegraphics[width=40mm, height=35mm]{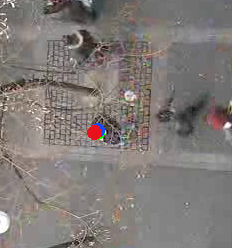}}
\subfigure{\label{fig:a23}\includegraphics[width=40mm, height=35mm]{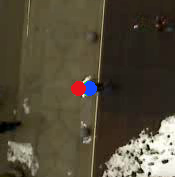}}
\subfigure{\label{fig:a24}\includegraphics[width=40mm, height=35mm]{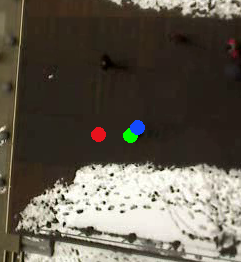}}\\
\vspace{-0.5em}
\subfigure{\label{fig:a31}\includegraphics[width=40mm, height=35mm]{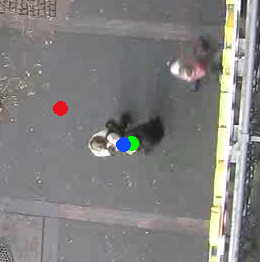}}
\subfigure{\label{fig:a32}\includegraphics[width=40mm, height=35mm]{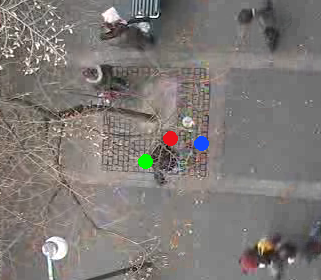}}
\subfigure{\label{fig:a33}\includegraphics[width=40mm, height=35mm]{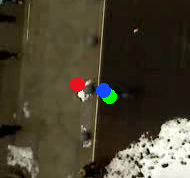}}
\subfigure{\label{fig:a34}\includegraphics[width=40mm, height=35mm]{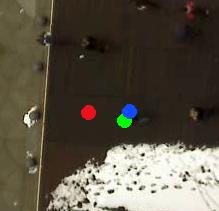}}\\
\vspace{-0.5em}
\subfigure{\label{fig:a41}\includegraphics[width=40mm, height=35mm]{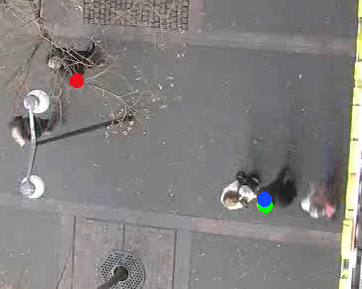}}
\subfigure{\label{fig:a42}\includegraphics[width=40mm, height=35mm]{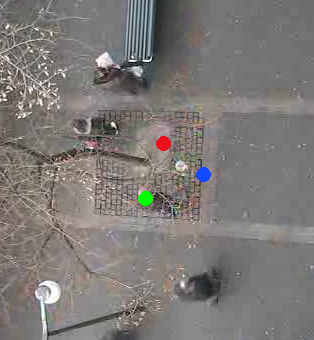}}
\subfigure{\label{fig:a43}\includegraphics[width=40mm, height=35mm]{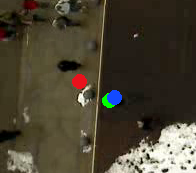}}
\subfigure{\label{fig:a44}\includegraphics[width=40mm, height=35mm]{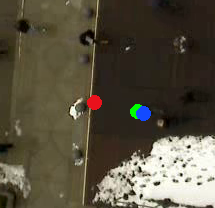}}\\
\vspace{-0.5em}
\subfigure{\label{fig:a51}\includegraphics[width=40mm, height=35mm]{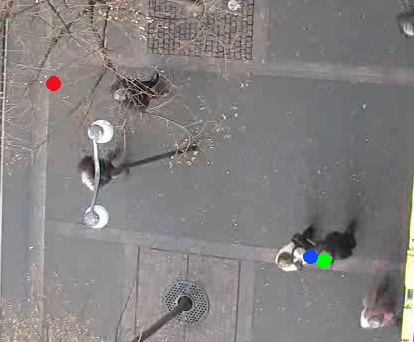}}
\subfigure{\label{fig:a52}\includegraphics[width=40mm, height=35mm]{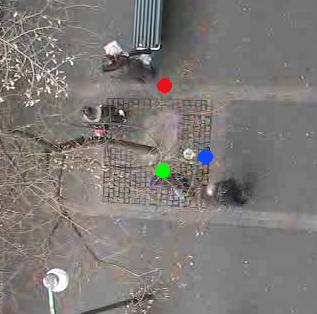}}
\subfigure{\label{fig:a53}\includegraphics[width=40mm, height=35mm]{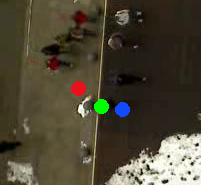}}
\subfigure{\label{fig:a54}\includegraphics[width=40mm, height=35mm]{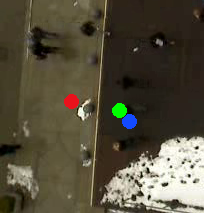}}
\caption{Temporal sequences visualization for different tracks drawn from both HOTEL and ETH dataset. The circles represent ground-truth (green), SNS-LSTM model (blue) and S-LSTM model (red), respectively. For each column, the first image shows the observed path (in green) which corresponds to 8 frames (the three circles are superimposed), while the remaining ones show the $9^{th}$, $13^{th}$, $17^{th}$ and $20^{th}$ predicted frames, respectively. In case of significant accelerations ($1^{st}$ column) our SNS-LSTM model remains close to the ground-truth compared to S-LSTM baseline avoiding obstacles and other pedestrians. By contrast, anomaly behaviors, i.e.. stopping after accelerating,  ($2^{nd}$ column) are not correctly predicted by both models. Semantic pooling mechanism avoids our model to collide with obstacles ($3^{th}$ column). Relying mainly on both social and navigation pooling mechanisms, our model predicts better positions than the S-LSTM model ($4^{th}$ column). Images are cropped to focus on monitored pedestrians' neighborhood.}\label{fig:seq_temporal}
\end{figure*}

\paragraph{Temporal visualization.} To temporally evaluate our predictions, we firstly visualize the observed path (\eg{}, $8$ frames) and then select four successive frames, namely $9^{th}, 13^{th}, 17^{th}$ and $19^{th}$ frame, of each trajectory. For the sake of simplicity, we only report results for SNS-LSTM and S-LSTM models, and ground-truths. More specifically, Fig.~\ref{fig:seq_temporal} shows observed paths and multiple predicted positions at successive timestamps for trajectories drawn from both HOTEL and ETH datasets, respectively. First column demonstrates the effect of the navigation pooling mechanism which avoids deviations from common paths and, at the same time, social pooling mechanism avoids collisions with nearby pedestrians. Second column shows an example of anomaly trajectories which do not frequently occur in the dataset, such as people suddenly stopping after accelerating. Such cases are not properly modelled by both models since predictions tend to move away from the ground-truths. Third column reports the effect of the semantic pooling mechanism which avoids the collision with an object (obstacle surrounded by snow) and predicts more realistic positions. In the last column, our SNS-LSTM model predicts more accurate positions than the S-LSTM model due to the combined effect of navigation and semantic pooling mechanisms.

The above temporal visualization confirms the effectiveness of the introduced elements to better capture complex human behaviors in crowded scenarios where also static objects contribute to the path generation process.
\section{Conclusion}
\label{sec:conclusion}
We proposed a comprehensive framework to model human-human and human-space interactions for trajectory forecasting in challenging scenarios. The SNS-LSTM merges past observations about the scene and semantics of crossed areas using Navigation and Semantic pooling mechanisms. Such an approach favours more reliable predicted paths when multiple choices are simultaneously possible to reach desired destinations. Previously observed paths and semantic labels of nearby cells allow a pedestrian to be aware of surrounding space and change his/her direction accordingly.

Experimental results show better performance than state-of-the-art methods which do not use context information. Our method, indeed, significantly reduce the error for common metrics and, from qualitative point of view, shows direction changes even when a pedestrian is not influenced by other humans in its neighborhood.
Our future work will investigate different datasets (\eg, Stanford Drone Dataset~\cite{Robicquet}) where more complex dynamics are captured as well as multiple agents settings since several elements, such as cars, cyclists or skateboarders, typically share the same environment. 

\paragraph*{Acknowledgements.}
This work is partially supported by MIUR (PRIN-2017 PREVUE grant) and UNIPD (BIRD-SID-2018 grant). We gratefully acknowledge the support of NVIDIA for their donation of GPUs used in this research.

{\small
\bibliographystyle{ieee}
\bibliography{mybib}
}

\end{document}